\documentclass{article}
\usepackage{spconf,amsmath,graphicx}
\usepackage {colortbl,array,xcolor}
\newcommand{\bhline}[1]{\noalign{\hrule height #1}}  

\newcolumntype{C}{>{\centering\arraybackslash}X}
\usepackage{booktabs}
\usepackage{multirow}
\usepackage{bm}
\usepackage{wrapfig}
\usepackage{url}
\usepackage{tabularx}

\title{PROMPT-FREE AND EFFICIENT SAM2 ADAPTATION FOR BIOMEDICAL SEMANTIC SEGMENTATION VIA DUAL ADAPTERS}
\name{Hinako Mitsuoka$^{\dagger}$ \qquad Kazuhiro Hotta$^{\ddagger}$
}
\address{Meijo University, 1-501 Shiogamaguchi, Tempaku-ku, Nagoya, 468-8502, Japan \\
$^{\dagger}$263441509@ccmailg.meijo-u.ac.jp \qquad $^{\ddagger}$kazuhotta@meijo-u.ac.jp
}
\begin{document}
%
\maketitle
\begin{abstract}
Segment Anything Model 2 (SAM2) demonstrated impressive zero-shot capabilities on natural images but faces challenges in biomedical segmentation due to significant domain shifts and prompt dependency.
To address these limitations, we propose a prompt-free, parameter-efficient fine-tuning framework designed for multi-class segmentation on variable-sized inputs. 
We introduce a convolutional Positional Encoding Generator to adapt effectively to arbitrary aspect ratios and present a dual-adapter strategy: High-Performance Adapter utilizing deformable convolutions for precise boundary modeling and Lightweight Adapter employing structural re-parameterization to minimize inference latency. 
Experiments on ISBI 2012, Kvasir-SEG, Synapse, and ACDC datasets demonstrate that our approach significantly outperforms strong adaptation baselines. 
Specifically, our method improved segmentation accuracy by up to 19.66\% over the vanilla SAM2, while reducing computational costs by approximately 87\% compared to heavyweight medical SAM adaptations, establishing a superior trade-off between accuracy and efficiency.
\end{abstract}

\begin{figure}[t!]
\begin{center}
\includegraphics[scale=0.45]{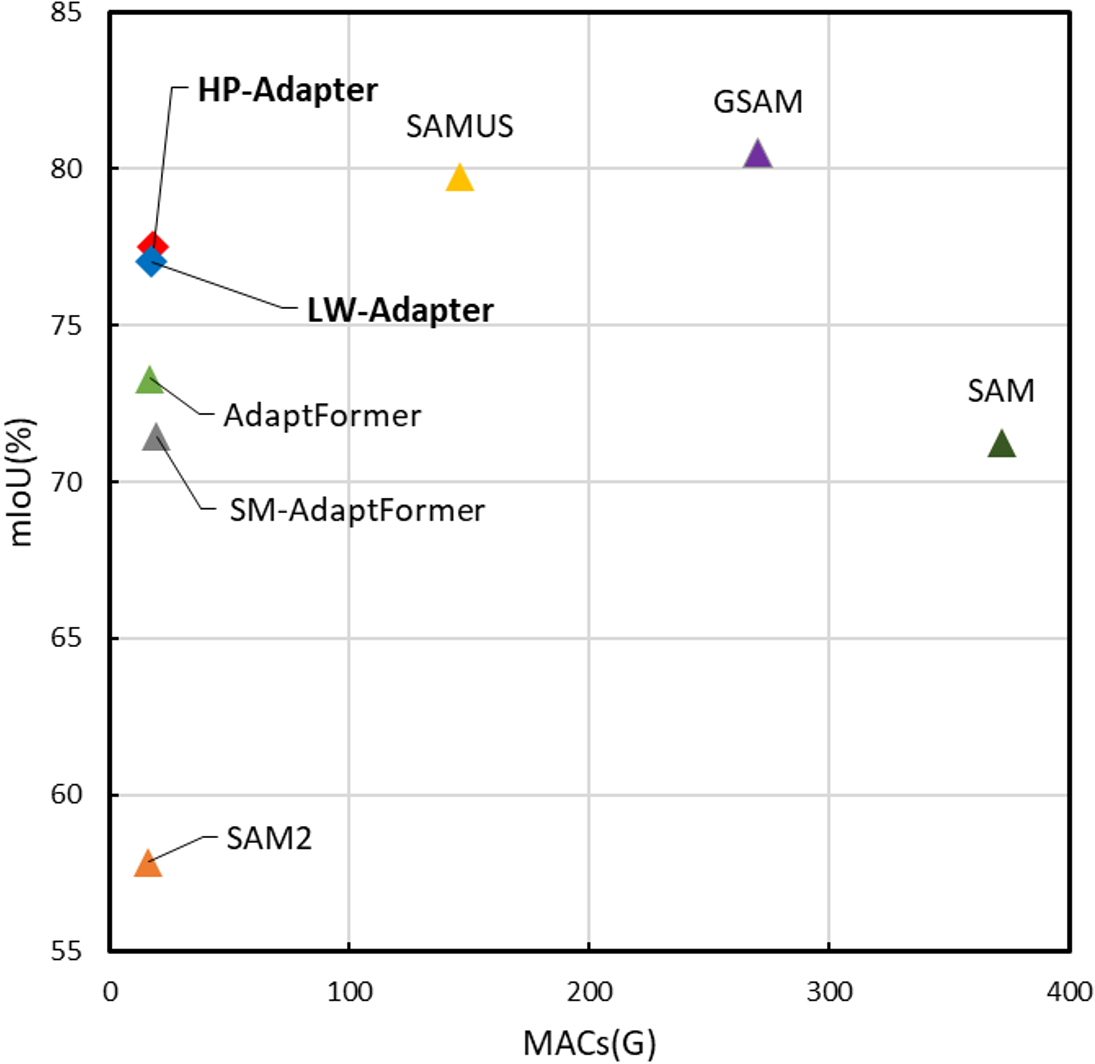}
\end{center}
\caption{Trade-off between mIoU and MACs for conventional PEFT methods and ours on the ISBI2012 dataset \cite{isbi}. The red and blue rhombi indicate our proposed Adapters, and the triangles indicate conventional methods. 
Note that SAM \cite{sam} and SAMUS \cite{samus} require fixed-length inputs. Therefore, their MACs are reported under their default input sizes, whereas the other methods are evaluated with variable-size inputs. In contrast, although SAM2 supports variable input sizes, it suffers from performance degradation due to the resolution mismatch ($1024^2$ pre-training vs.\ $256^2$ inference), where the interpolation of absolute positional embeddings disrupts feature representation.
Our approach effectively bridges this gap, achieving a superior trade-off.
}
\label{fig:tr}
\end{figure}

%
\section{Introduction}
\label{sec:intro}

Biomedical semantic segmentation is essential for computer-aided diagnosis and quantitative analysis.
Although specialized models for semantic segmentation, such as U-Net \cite{unet}, perform well, they are typically trained from scratch per dataset.
Recently, foundation models such as SAM \cite{sam} and SAM2 \cite{sam2} have shown strong generalization via large-scale pre-training, and SAM2 further enables fast inference with the Image Encoder using Hiera \cite{hiera}.

However, adapting SAM2 to biomedical images remains non-trivial due to domain shift (e.g., low contrast and complex structures),  prompt dependency that hinders fully automatic pipelines, and variable input sizes in medical imaging, though SAM-style models are often used with fixed input resolutions for stable performance.
Parameter-efficient fine-tuning (PEFT) methods, such as SAMUS \cite{samus} and GSAM \cite{gsam}, alleviate some issues but may introduce extra components that increase inference overhead.

We propose a prompt-free, PEFT framework for SAM2 that targets a favorable mIoU-MACs trade-off illustrated in Fig.\ref{fig:tr}.
We freeze the prompt encoder and extend the mask decoder to output class-specific masks in a single forward pass.
To robustly handle arbitrary resolutions, we replace absolute positional embeddings with a convolutional Positional Encoding Generator (PEG) \cite{peg}.
For encoder adaptation, we introduce two plug-in adapters: High-Performance (HP) Adapter with DCNv2 \cite{dcnv2} for higher accuracy and Lightweight (LW) adapter that uses structural re-parameterization \cite{repvgg} to collapse multi-branch convolutions at inference.

Experiments on four datasets: ISBI2012 \cite{isbi}, Kvasir-SEG \cite{kvasir}, Synapse \cite{smo}, and ACDC \cite{acdc}, showed that our adapters consistently improved SAM2, with up to +19.66\% mIoU over vanilla SAM2 while incurring only a small increase in inference-time MACs.


Our main contributions can be summarized as follows.
\begin{itemize}
    \setlength{\itemsep}{0pt}
    \item Prompt-free SAM2 adaptation for multi-class segmentation on variable-sized biomedical images.
    \item Dual adapters, HP and LW-Adapter, offering an explicit accuracy-efficiency trade-off.
    \item Consistent gains on ISBI2012, Kvasir-SEG, Synapse, and ACDC with low inference overhead.
\end{itemize}

\section{Related Works}
\label{sec:related}

\subsection{Parameter Efficient Fine-Tuning (PEFT)}
\label{sub:peft}

Parameter Efficient Fine-Tuning (PEFT) adapts large pre-trained models by updating only a small subset of parameters while keeping most weights frozen, reducing training cost and storage.
Representative approaches include low-rank adaptation, such as LoRA \cite{lora} and adapter-based tuning that inserts lightweight modules into Transformer blocks (e.g., AdaptFormer \cite{adaptformer}).
In this work, we focus on PEFT designs that preserve inference efficiency, targeting a favorable accuracy--MACs trade-off.

\subsection{Segmentation and Foundation Models}
\label{sub:seg}
Since U-Net \cite{unet} established a strong baseline for medical image segmentation, various architectures \cite{deeplab} have been developed.
However, such task-specific models often require substantial training from scratch.

Recently, Segment Anything Model (SAM) \cite{sam} demonstrated strong generalization via large-scale pre-training.
SAM2 adopts a hierarchical Hiera \cite{hiera} encoder and a memory mechanism for video segmentation, enabling multi-scale processing and fast inference.
Despite these advances, applying SAM2 directly to biomedical segmentation remains challenging due to domain shift and the lack of domain-specific semantic understanding required for automatic segmentation.
We therefore leverage SAM2 as a foundation and adapt it to biomedical domains using lightweight modules.

\subsection{Fine-tuning of SAM}
\label{sub:ftsam}

Full fine-tuning of SAM or SAM2 is computationally expensive; thus, efficient adaptation has attracted increasing attention.
LoRA \cite{lora} has been widely used for SAM-style models, and variants such as Conv-LoRA \cite{convlora} incorporate convolutions to strengthen local feature modeling.
Adapter-based methods such as AdaptFormer \cite{adaptformer} provide another efficient alternative by inserting compact modules into Transformer feed-forward layers.

Regarding variable input sizes, SAMUS \cite{samus} uses CNN-Transformer fusion, and GSAM \cite{gsam} supports variable resolutions via an auxiliary CNN encoder.
However, relying on external encoders can increase inference overhead.
In contrast, we extend efficient adaptation to SAM2 with internal, lightweight adapters and convolutional positional encoding, explicitly minimizing additional MACs while improving boundary-sensitive biomedical segmentation.

\section{Methodology}
\label{sec:method}

\begin{figure}[t]
\begin{center}
\includegraphics[scale=0.32]{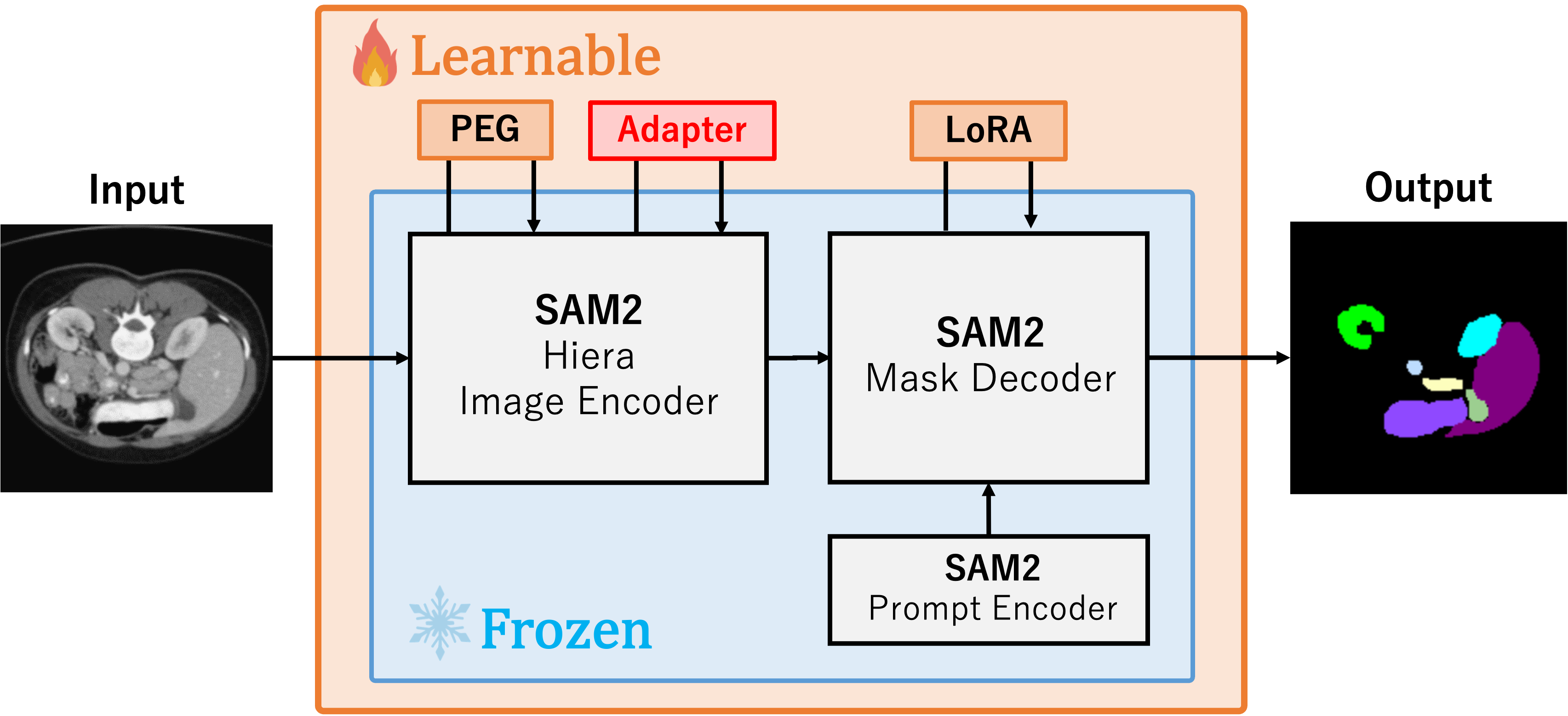}
\end{center}
\caption{Overview of the proposed method, SAM2 fine-tuning framework. $Frozen$ indicates a network in which the weight parameters are fixed, and $Learnable$ indicates a network in which the weight parameters are updated.}
\label{fig:teian}
\end{figure}

Fig.\ref{fig:teian} illustrates the overview of our method. 
During fine-tuning, we keep the original SAM2 components (Image Encoder, Prompt Encoder, and Mask Decoder) frozen, and update only the newly introduced modules marked as $Learnable$.

\subsection{Prompt-Free Multi-Class Segmentation With SAM2}

When still images are fed into SAM2, SAM2 follows the SAM-style modular design with Image Encoder, Prompt Encoder, and Mask Decoder.
Given an image and a user prompt (e.g., point/box/mask), the decoder predicts the corresponding object mask.
Although this prompt-conditioned formulation is flexible for interactive use and larger pipelines, it is not ideal for fully automatic semantic segmentation.

Inspired by prior work that automates prompt-based SAM models for downstream segmentation \cite{convlora, gsam}, we remove the need for user prompts by freezing the prompt encoder and feeding a fixed set of prompt tokens to the mask decoder for all samples.
Concretely, during fine-tuning and inference, the prompt input is replaced with pre-trained constant tokens $\mathbf{T}$,
yielding a prompt-free forward pass while preserving the original decoder interface.
The original mask decoder is designed for prompt-conditioned binary mask prediction (foreground vs.\ background).
To support multi-class semantic segmentation with $C$ classes in a single forward pass, we modify the decoder outputs to produce class-specific masks directly.
Specifically, we allocate $C$ output tokens and use a per-class hypernetwork to map each token to a dense mask logit map, resulting in $C$ mask logits $\{\mathbf{m}_c\}_{c=1}^{C}$ without an explicit classification branch or score aggregation.
This enables end-to-end multi-class segmentation without any user interaction.

We adapt the mask decoder in a parameter-efficient manner by applying LoRA to its transformer layers while keeping the original mask-decoder weights frozen; only the LoRA parameters are updated during fine-tuning.
To handle variable input sizes robustly, we replace absolute positional embeddings with a convolutional Positional Encoding Generator (PEG) \cite{peg}.
Encoder-side adaptation is performed via lightweight plug-in adapters, which we describe next.

\subsection{High-Performance Adapter}

We propose two adapter variants for parameter-efficient adaptation and instantiate them in SAM2.
Both adapters follow the AdaptFormer design \cite{adaptformer} and are inserted in parallel to the MLP branch of each transformer block in the Image Encoder.

\begin{figure}[t]
\begin{center}
\includegraphics[scale=0.4]{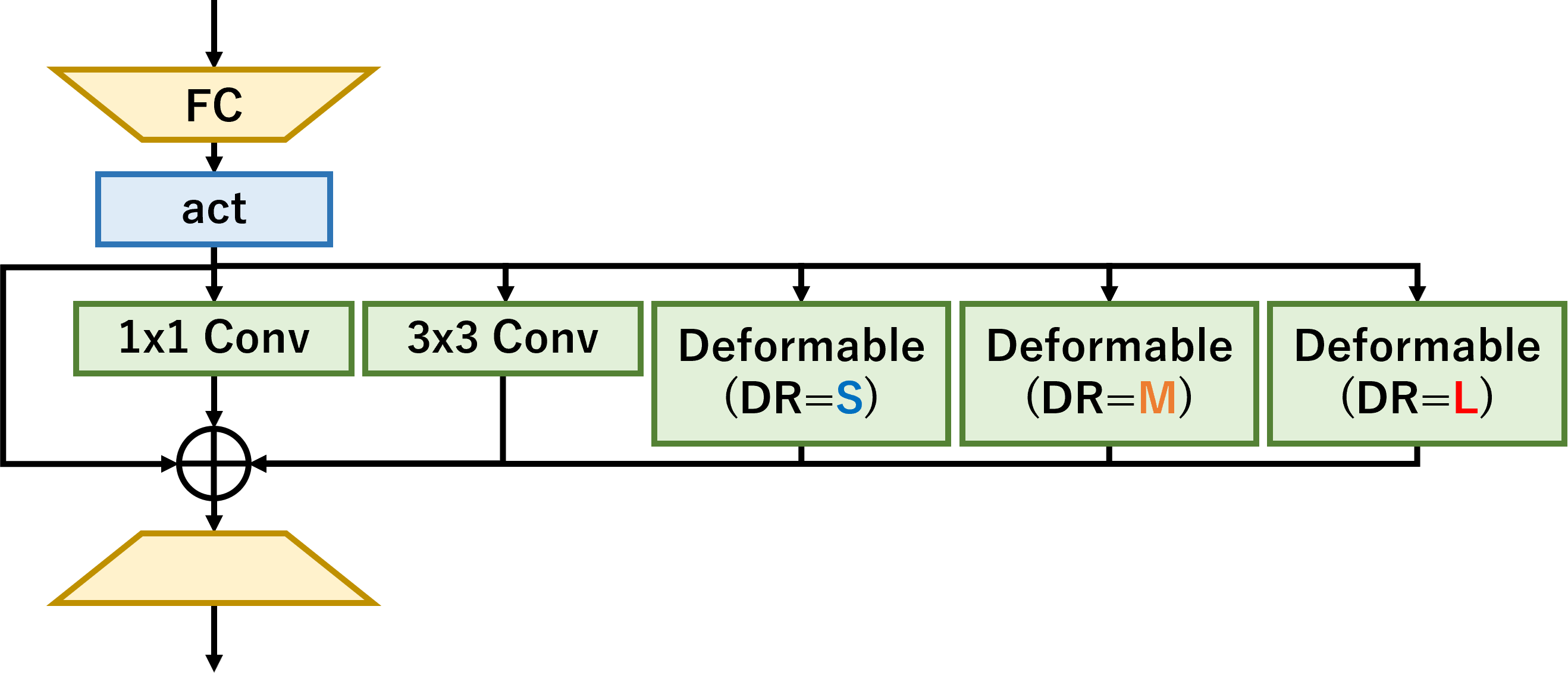}
\end{center}
\caption{Overview of High-Performance (HP) Adapter. DCNv2 layers are inserted between FC layers to enhance boundary modeling. "act" is GELU \cite{gelu}.}
\label{fig:hp}
\end{figure}

Fig.\ref{fig:hp} shows the High-Performance (HP) Adapter.
It places three Deformable Convolution v2 layers \cite{dcnv2} between two fully connected layers, together with skip connections and lightweight $1\times1$ and $3\times3$ convolutions.
To enlarge the effective receptive field when predicting offsets and modulation masks, we apply a dilation rate (denoted as DR in Fig.\ref{fig:hp}).
We use DR $\in \{12, 24, 36\}$ (labeled as S/M/L) selected on a validation split.
This design enables boundary-aware adaptation by combining local details with broader contextual cues, leveraging the adaptive sampling of Deformable Convolution v2 to capture the irregular geometries and non-rigid deformations typical in biomedical images.

\subsection{Lightweight Adapter using Re-parameterization}

\begin{figure}[t]
\begin{center}
\includegraphics[scale=0.4]{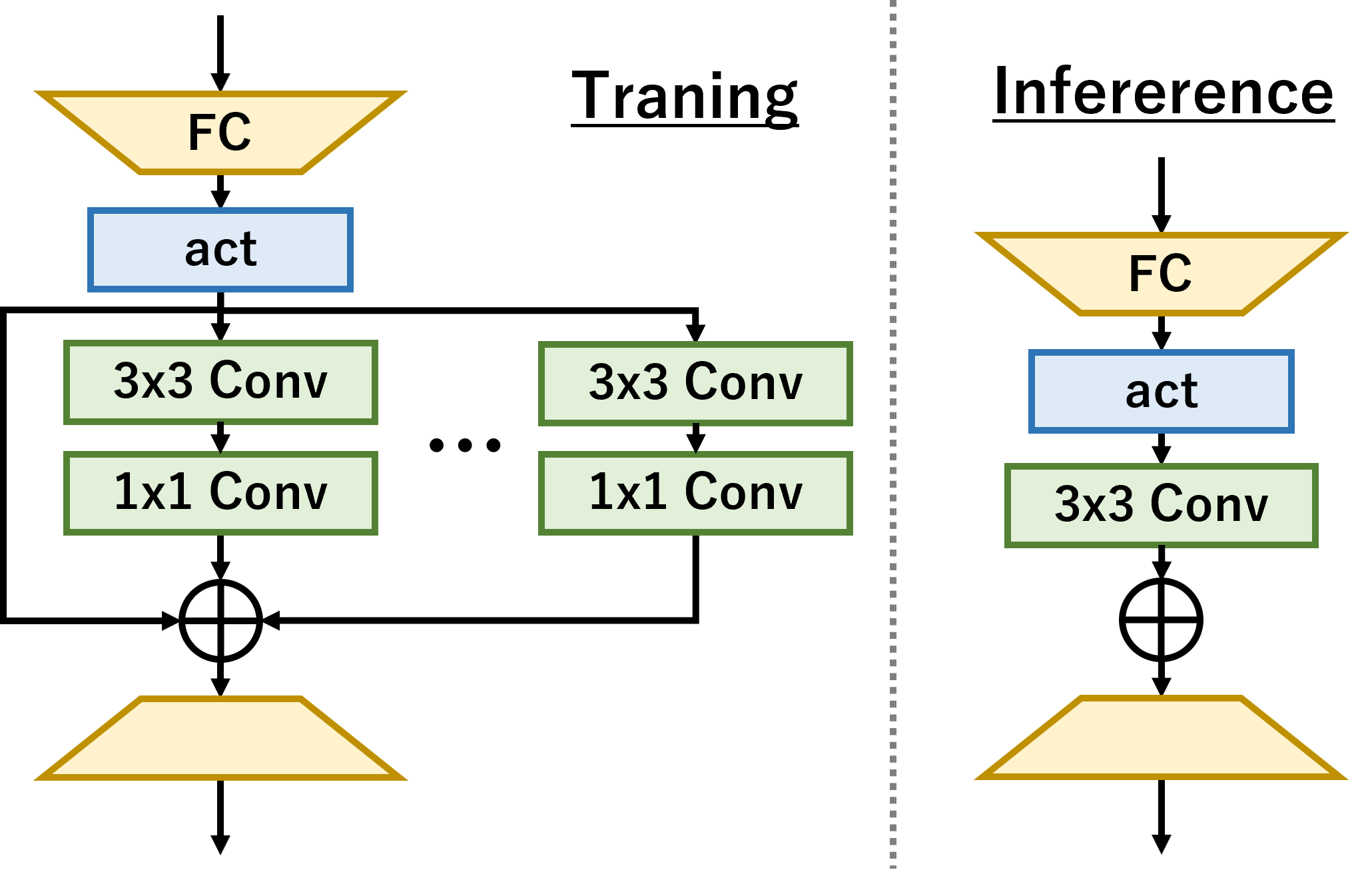}
\end{center}
\caption{Overview of Lightweight (LW) Adapter using Re-parameterization. The multi-branch conv block during training is fused into a $3\times3$ conv at inference. "act" is GELU.}
\label{fig:lw}
\end{figure}

Fig.\ref{fig:lw} shows the Lightweight (LW) Adapter.
Drawing inspiration from \cite{gcnet}, which achieves real-time performance through structural re-parameterization, we design this adapter to decouple training dynamics from inference latency.
Specifically, between two fully connected layers, we adopt a multi-branch design consisting of an identity branch and multiple convolutional branches formed by stacked $3 \times 3$ and $1 \times 1$ convolutions.
We set the number of convolutional branches to $3$ based on validation experiments.

During training, this multi-branch structure facilitates gradient flow and enables the model to learn richer representations.
At inference, we employ structural re-parameterization \cite{repvgg} to mathematically fuse these branches into a single equivalent $3 \times 3$ convolution.
Although recent methods such as RepAdapter \cite{luo2023towards} utilize re-parameterization to merge adapters into the backbone weights for zero-cost adaptation, our approach distinctively focuses on the internal topology of the adapter itself. 
By leveraging a multi-branch design during training and collapsing it at inference, we maximize the representational capacity of the adapter without incurring the computational penalty of complex branches.

\section{Experiments}
\label{sec:experiment}

\begin{table*}[t!]
    \centering
    \caption{Comparison results on four datasets. Random cropping is applied to approaches except for SAM. The size of input images is $1024\times1024$ for SAM, $256\times256$ for SAMUS, and "Image size" in this table for the other comparison methods. \textbf{Bold} indicates the best results among SAM2-based methods.}
    \label{tab:result}
    \scalebox{0.9}{
    \begin{tabularx}{15cm}{rCCCC}
    \bhline{1.5pt}
    \multicolumn{1}{r}{Dataset} & \multicolumn{1}{C}{ISBI2012 \cite{isbi}} & \multicolumn{1}{C}{Kvasir-SEG \cite{kvasir}} & \multicolumn{1}{C}{Synapse \cite{smo}}& \multicolumn{1}{C}{ACDC \cite{acdc}}\\
    \hline
    \multicolumn{1}{r}{Class} & \multicolumn{1}{C}{2} & \multicolumn{1}{C}{2} & \multicolumn{1}{C}{9} & \multicolumn{1}{C}{4}\\
    \hline
    \multicolumn{1}{r}{Image size} & \multicolumn{1}{C}{$256 \times 256$} & \multicolumn{1}{C}{$224 \times 224$}& \multicolumn{1}{C}{$224 \times 224$}& \multicolumn{1}{C}{$256 \times 256$}\\
    \hline 
    \multirow{2}{*}{$\rm{Random\ crop}^ \ast$} & \multirow{2}{*}{$128 \times 128$} & $224 \times 224$& $224 \times 224$& \multirow{2}{*}{$128 \times 128$}\\
    &&+ padding&+ padding&\\
    \hline
    \rowcolor[gray]{0.9}
    U-Net \cite{unet}
    &80.39&86.58&69.93&82.53\\ 
    \hline
    \textit{SAM-based Method} \\
    \hline
    \rowcolor[gray]{0.9}
    SAM \cite{sam}
    &71.27&79.36&40.61&84.16\\
    \rowcolor[gray]{0.9}
    SAMUS \cite{samus}
    &79.75&90.13&77.53&87.29\\
    \rowcolor[gray]{0.9}
    GSAM \cite{gsam}
    &80.52&86.99&72.88&87.83\\
    \hline
    \textit{SAM2-based Method} \\
    \hline
    SAM2 \cite{sam2}
    &57.85&82.99&61.11&81.14\\
    AdaptFormer \cite{adaptformer}
    &73.30&72.49&54.50&62.41 \\
    SM-AdaptFormer \cite{gsam}
    &71.44&80.48&64.29&83.25 \\
    \hline 
    Ours (HP-Adapter)
    &\textbf{77.51}&\textbf{85.84}&\textbf{66.62}&\textbf{85.20}\\
    Ours (LW-Adapter)
    &77.04&84.75&64.72&85.00\\
    \bhline{1.5pt}
    \end{tabularx}
    }
\end{table*}

\subsection{Datasets and Metrics}
\label{sub:dataset}

In our experiments, we focus on biomedical image segmentation across multiple modalities and input resolutions. 
Specifically, we used the ISBI2012 dataset \cite{isbi} (2 classes) as a cell microscopy image, and three medical imaging datasets, Kvasir-SEG \cite{kvasir} (2 classes) for endoscopic images, Synapse multi-organ dataset \cite{smo} (Synapse, 9 classes) for CT, and ACDC \cite{acdc} (4 classes) for cardiac MRI.
The pixel counts for each dataset are listed in Tab.\ref{tab:result}.

We adopt Intersection over Union (IoU) as the primary metric for semantic segmentation, where IoU quantifies the overlap between predictions and ground truth. We report mean IoU (mIoU), defined as the class-wise IoU averaged across all classes.

\subsection{Training Conditions}
\label{sub:implementation}

In this paper, we utilized the PyTorch library and trained the model using the Adam optimizer for 200 epochs with a batch size of 2.
The learning rate was initially set to 0.005 and gradually decreased using the cosine scheduler.
For comparison, we evaluate a conventional CNN-based segmentation baseline, U-Net \cite{unet}. 
We also compare against SAM and representative parameter-efficient adaptation methods for SAM, including SAMUS \cite{samus} and GSAM \cite{gsam}. 
To assess encoder-side tuning for SAM2, we consider AdaptFormer \cite{adaptformer} and SM-AdaptFormer introduced in GSAM \cite{gsam}, which augments AdaptFormer with parallel dilated convolutional branches to introduce multi-scale spatial features.
Finally, we report results with our proposed two adapter variants for SAM2, enabling an explicit trade-off between accuracy and MACs.
For SAM and SAM2, we freeze the Image and Prompt Encoders and fine-tune only the Mask Decoder.
For adapter-based baselines, we change only the adapter modules under the same trainable-parameter setting as ours, while SAMUS and GSAM follow their original training protocols.

For data augmentation in training, we use random cropping, horizontal flipping, and random rotation for all methods that can ingest variable-sized inputs (i.e., all approaches excluding SAM, which requires a fixed input size: $1024\times1024$). 
The crop sizes used for random cropping are listed in Tab.\ref{tab:result}.

\subsection{Quantitative Results.}
\label{sub:quantitative}

Tab.\ref{tab:result} reports quantitative results on four biomedical segmentation datasets in terms of mIoU (\%).
We compare conventional CNN-based segmentation (U-Net), SAM variants (SAM, SAMUS, GSAM), and PEFT baselines on SAM2 (AdaptFormer and SM-AdaptFormer), against our two adapter designs.

Overall, our methods consistently improved the performance of SAM2 across all datasets.
In particular, the proposed HP-Adapter achieved the best accuracy among SAM2-based approaches, reaching 77.51\% on ISBI2012, 85.84\% on Kvasir-SEG, 66.62\% on Synapse, and 85.20\% on ACDC.
Compared with SAM2, HP-Adapter improved the accuracy by +19.66, +2.85, +5.51, and +4.06 on all datasets, respectively.
LW-Adapter also improved SAM2 while prioritizing efficiency in inference.

Among PEFT baselines, AdaptFormer showed limited transferability on biomedical datasets, while SM-AdaptFormer performed better but remained below our adapters.
We hypothesize that, while SM-AdaptFormer enhances context via fixed dilated convolutions, its sampling pattern remains static. In biomedical images with large shape variability and boundary ambiguity, deformation-aware modeling (HP-Adapter) is more effective.
Although SAMUS and GSAM achieved strong performance, they rely on SAM-style fixed-resolution processing or additional components. 
In contrast, our approach built upon SAM2 and improved accuracy via lightweight adapters with minimal inference overhead.
These results demonstrated that the proposed adapters effectively bridge the domain gap for biomedical segmentation while maintaining an optimal balance between accurate segmentation and inference efficiency.

\subsection{Qualitative Results.}
\label{sub:qualitative}

\begin{figure}[t]
\begin{center}
\includegraphics[scale=0.36]{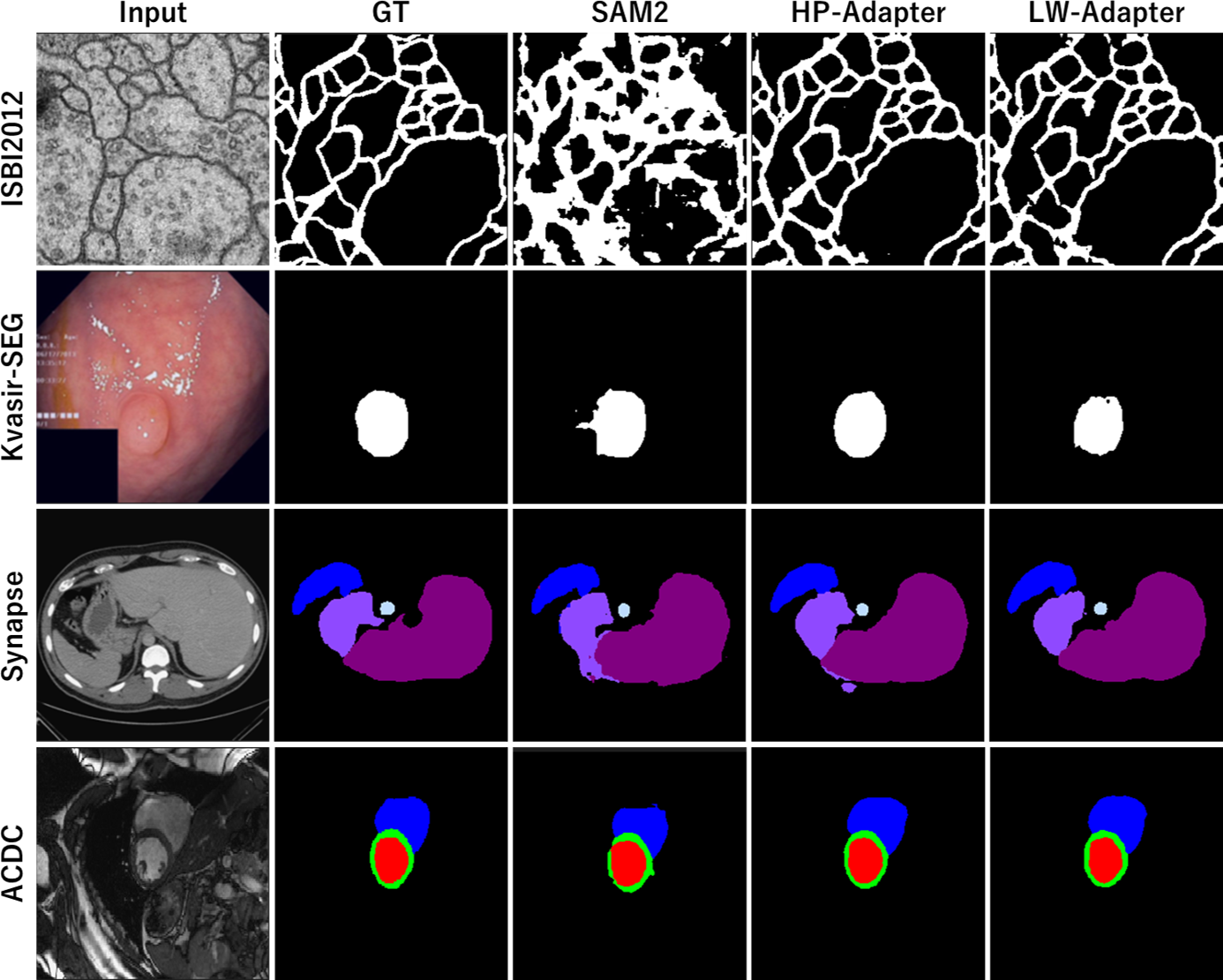}
\end{center}
\caption{Qualitative results. From left to right show Input image, Ground truth, SAM2 \cite{sam2}, HP-Adapter, LW-Adapter.}
\label{fig:kashika}
\end{figure}

Fig.\ref{fig:kashika} presents a visual comparison of segmentation results across representative samples.
As observed in the third column, the baseline SAM2 often struggles with the ambiguity of biomedical boundaries, leading to noticeable over-segmentation or failure to capture the complete cell or organ structure due to the significant domain gap.

In contrast, our proposed methods generate high-fidelity masks that closely align with the Ground Truth (GT).
Specifically, the HP-Adapter demonstrates superior capability in delineating complex and irregular boundaries. 
This improvement is attributed to Deformable Convolution v2, which effectively models geometric deformations inherent in organs.
Remarkably, our LW-Adapter produces segmentation masks that are visually comparable to those of the HP-Adapter, showing minimal degradation despite its lightweight architecture.
This qualitative evidence corroborates our quantitative findings, confirming that our re-parameterization strategy successfully preserves representational power while optimizing for inference efficiency.

\subsection{Efficiency Analysis.}
\label{sub:ablation}

Tab.\ref{tab:ablation-macs} and Fig.\ref{fig:tr} present a comprehensive comparison of inference computational cost (MACs), model size (Params), and segmentation accuracy (mIoU).
For a fair comparison, we measure MACs on the ISBI2012 dataset using an input resolution of $256\times256$ for all methods that accept this resolution, while SAM is evaluated at its required resolution of $1024\times1024$.
SAM achieved decent accuracy but incurred a massive computational cost of 371.98 GMACs due to its heavy encoder. 
Although specialized medical adaptations like SAMUS and GSAM achieved high accuracy, they remained computationally expensive due to their additional modules or external encoders, limiting their applicability in resource-constrained environments. 
On the other hand, the baseline SAM2, despite being extremely efficient, yielded a poor mIoU of 57.85\% due to the lack of domain adaptation.
Our proposed method successfully bridges this gap, achieving an optimal trade-off between accuracy and efficiency. 
Compared to SAMUS, our HP-Adapter achieved comparable accuracy while slashing the computational cost by approximately 87\%. 
Furthermore, our approach outperforms other parameter-efficient tuning methods of SAM2, such as AdaptFormer and SM-AdaptFormer, in terms of accuracy. 
Notably, our LW-Adapter further minimizes the computational cost to 17.58 GMACs while maintaining a high mIoU of 77.04\%, demonstrating the effectiveness of our re-parameterization strategy in delivering robust biomedical segmentation with significantly lower latency.

\begin{table}[t]
    \centering
    \caption{Comparison of MACs, parameters, and segmentation accuracy on the ISBI2012 dataset. 
    Note that SAM and SAMUS use fixed input resolutions, while other methods allow variable inputs. \textbf{Bold} indicates the best results among existing methods, while our proposed methods are listed in the bottom section.
    }
    \scalebox{0.92}{
    \begin{tabular*}{9cm}{@{\extracolsep{\fill}}lccc}
    \bhline{1.5pt}
    \multicolumn{1}{l}{Methods} & \multicolumn{1}{c}{MACs(G)} & \multicolumn{1}{c}{Params(M)} & \multicolumn{1}{c}{mIoU} \\      
    \hline
        SAM \cite{sam} &371.98&90.49&71.27 \\
        SAMUS \cite{samus} &145.87&130.10&79.75 \\
        GSAM \cite{gsam} &270.33&168.35&\textbf{80.52} \\
        SAM2 \cite{sam2} &\textbf{16.27}&\textbf{73.43}&57.85 \\
        AdaptFormer \cite{adaptformer} &16.89&76.34&73.30 \\
        SM-AdaptFormer \cite{gsam} &19.67&89.74&71.44 \\
        \hline
        \textbf{Ours (HP-Adapter)} &18.24&81.93&77.51 \\
        \textbf{Ours (LW-Adapter)} &17.58&79.61&77.04 \\
    \bhline{1.5pt}
    \end{tabular*}
    \label{tab:ablation-macs}
    }
\end{table}

\section{Conclusion}
\label{sec:conclusion}

In this paper, we presented a prompt-free, PEFT framework that adapts SAM2 for fully automatic biomedical image segmentation.
By integrating PEG and replacing the positional encoding method of SAM2, our approach robustly handles the diverse resolutions and aspect ratios inherent in medical imaging without user intervention.
Our core contribution lies in the dual-adapter design: the HP Adapter leverages Deformable Convolutions to capture intricate boundary details, while the LW Adapter employs structural re-parameterization to ensure minimal inference latency.
Experiments on ISBI2012, Kvasir-SEG, Synapse, and ACDC demonstrated consistent gains over strong SAM/SAM2 adaptation baselines, including up to +19.66 mIoU over vanilla SAM2, while reducing computational costs by approximately 87\% compared to heavyweight state-of-the-art adaptations like SAMUS.

\bibliographystyle{IEEEbib}
\bibliography{strings,refs}

\end{document}